\pdfoutput=1

\documentclass[11pt]{article}

\usepackage[final]{acl}

\usepackage{times}
\usepackage{latexsym}
\usepackage{amsmath}

\usepackage[T1]{fontenc}

\usepackage[utf8]{inputenc}

\usepackage{microtype}

\usepackage{inconsolata}

\usepackage{graphicx}
\usepackage{xspace}
\usepackage{animate}
\usepackage{makecell}
\usepackage[ruled,vlined]{algorithm2e}
\usepackage{tikz}
\usepackage{subcaption}
\usepackage{xcolor}
\newcommand\method{\text{AnnaAgent}\xspace}
\newcommand{\midsmall}{\fontsize{10.5pt}{12pt}\selectfont}

\title{AnnaAgent: Dynamic Evolution Agent System with Multi-Session Memory for Realistic Seeker Simulation}

\author{
 \textbf{Ming Wang\textsuperscript{1}},
 \textbf{Peidong Wang\textsuperscript{1}},
 \textbf{Lin Wu\textsuperscript{2}},
 \textbf{Xiaocui Yang\textsuperscript{1}},
\\
 \textbf{Daling Wang\textsuperscript{1,}\thanks{Corresponding author.}},
 \textbf{Shi Feng\textsuperscript{1}},
 \textbf{Yuxin Chen\textsuperscript{3}},
 \textbf{Bixuan Wang\textsuperscript{2,4}},
 \textbf{Yifei Zhang\textsuperscript{1}}
\\
\\
 \textsuperscript{1}School of Computer Science and Engineering, Northeastern University,
 \\
 \textsuperscript{2}Mental Health Education Center, Northeastern University,
 \\
 \textsuperscript{3}School of Sociology and Psychology, Central University of Finance and Economics,
 \\
 \textsuperscript{4}School of Psychology, Northeast Normal University,
\\
 \midsmall{
   \href{mailto:sci.m.wang@gmail.com}{sci.m.wang@gmail.com},
   \href{mailto:wulin@mail.neu.edu.cn}{wulin@mail.neu.edu.cn},
   \href{mailto:pdongwang@163.com}{pdongwang@163.com},}\\
\midsmall{\{yangxiaocui,wangdaling,fengshi,zhangyifei\}@cse.neu.edu.cn,}\\
\midsmall{
   \href{mailto:2024212301@email.cufe.edu.cn}{2024212301@email.cufe.edu.cn},
   \href{mailto:wangbixuan@nenu.edu.cn}{wangbixuan@nenu.edu.cn}
   }
}

\begin{document}
\maketitle
\begin{abstract}
Constrained by the cost and ethical concerns of involving real seekers in AI-driven mental health, researchers develop LLM-based conversational agents (CAs) with tailored configurations, such as profiles, symptoms, and scenarios, to simulate seekers. While these efforts advance AI in mental health, achieving more realistic seeker simulation remains hindered by two key challenges: dynamic evolution and multi-session memory.
Seekers' mental states often fluctuate during counseling, which typically spans multiple sessions. To address this, we propose \textbf{\method}\footnote{The name comes from ``Anna O.'', the pseudonym for Bertha Pappenheim, a patient of Josef Breuer whose case significantly influenced the development of psychoanalysis. Her treatment marked the first use of the ``Talking Cure'', laying the foundation for modern psychotherapy and highlighting the importance of verbal expression in psychological healing.}, an emotional and cognitive dynamic agent system equipped with tertiary memory. \method incorporates an emotion modulator and a complaint elicitor trained on real counseling dialogues, enabling dynamic control of the simulator's configurations. Additionally, its tertiary memory mechanism effectively integrates short-term and long-term memory across sessions. Evaluation results demonstrate that \method achieves more realistic seeker simulation in psychological counseling compared to existing baselines. The ethically reviewed code can be found on \href{https://github.com/sci-m-wang/AnnaAgent}{https://github.com/sci-m-wang/AnnaAgent}.
\end{abstract}


\section{Introduction}
The issue of mental disorders is a critical challenge for ongoing society \cite{WHO2022world}. Effective psychological counseling plays a crucial role in addressing these challenges, yet the availability of trained therapists remains limited \cite{stewart2022global,APPG2021new}.
Researchers introduce AI to provide mental health support to alleviate the shortage of human counselors \cite{li_systematic_2023}. 
\begin{figure}[!ht]
    \centering
    \begin{subfigure}[t]{0.48\textwidth}
        \includegraphics[width=\linewidth]{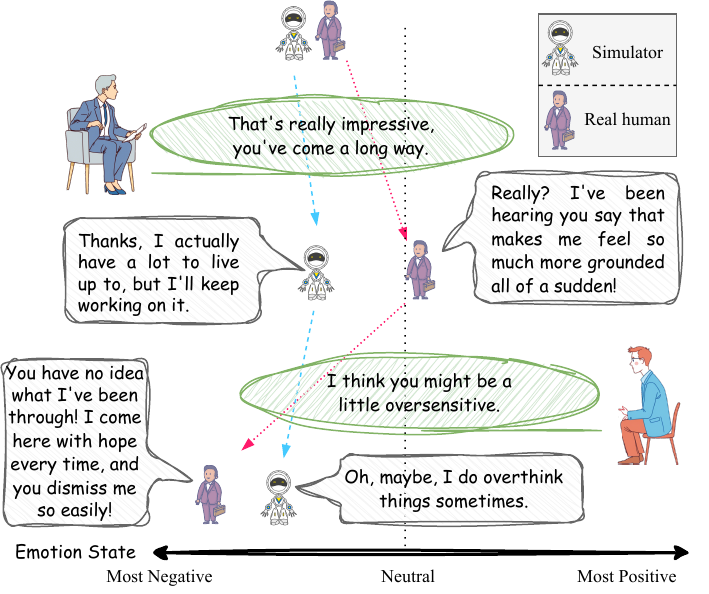}
        \vspace{-0.3cm}
        \caption{The challenge of dynamic evolution. The robot logo represents an LLM-based CA, a simulated seeker, while the man logo represents a real seeker. The \textcolor[HTML]{45CAFF}{blue} dashed line indicates the evolution of the simulator's emotional state, and the \textcolor[HTML]{ff1b6b}{red} dotted line indicates those of the real seeker. CAs' responses usually maintain stable emotions, whereas real seekers’ emotional fluctuations are more pronounced during counseling.}
        \label{fig:problem_a}
    \end{subfigure}
    \vspace{1em}
    \begin{subfigure}[t]{0.48\textwidth}
        \includegraphics[width=\linewidth]{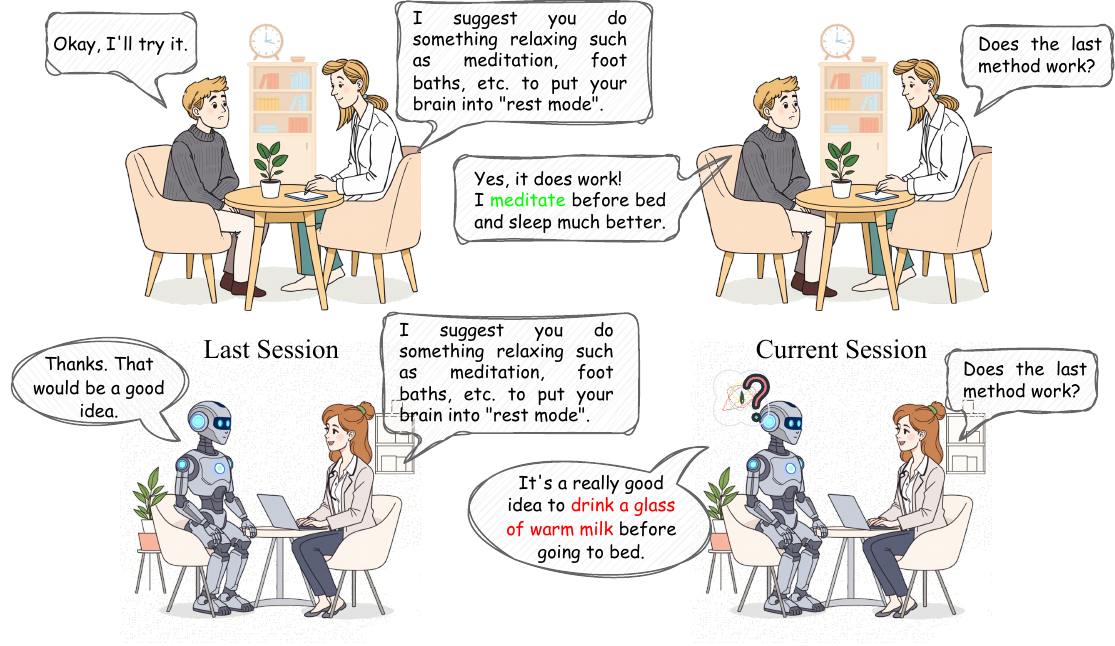}
        \vspace{-0.3cm}
        \caption{The challenge of multi-session memory. The left side indicates the previous session, while the right side indicates the current session. When it comes to the topic in the previous session, the responses of the real seeker contain correct information (indicated in \textcolor[HTML]{00ff00}{green}), while the simulator without memory contains incorrect responses (information in \textcolor[HTML]{ff0000}{red}).}
        \label{fig:problem_b}
    \end{subfigure}
    \vspace{-1.7em}
    \caption{Examples of the two challenges in seeker simulation. Subfigure (a) for dynamic evolution and (b) for multi-session memory.}
    \vspace{-1em}
    \label{fig:challenges}
\end{figure}
As one of the most important constituents of mental health research, seekers are crucial for data construction, effect evaluation, and ethical analyses. However, the introduction of large numbers of real seekers for research not only creates a high-cost burden but may also lead to ethical concerns. Thus, researchers design LLM-based conversational agents (CAs) to simulate seekers.
\citet{chen2023llmempoweredchatbotspsychiatristpatient} designed prompts for CAs based on the psychiatrist's goals of communicating with seekers. Similarly to them, \citet{duro_introducing_2024} designed specialized prompts and invited psychotherapy professionals to participate in the adaptation of the prompts. \citet{lan2024depressiondiagnosisdialoguesimulation} initialized a batch of seeker agents using seeker profiles generated by GPT-4 \cite{openai2024gpt4technicalreport} collected based on the D$^4$ dataset \cite{yao2022d4chinesedialoguedataset}. Accommodating different profiles, like seekers' personal backgrounds and symptom descriptions in prompts, is widely used to simulate the conversational behavior of real seekers.

Undeniably, these works have made a significant contribution to the study of seeker simulation, addressing issues like CAs' easily endorsing advice, sycophantic obedience \cite{wester_this_2024,black_-humanizing_2023}, and emotional flatness \cite{wang_enhancing_2023,balcombe_ai_2023}.
However, CAs designed with existing methods are still not realistic enough, blocking the further development of AI for mental health research. As shown in Figure \ref{fig:problem_a}, A real seeker is usually more sensitive due to the psychological disorders suffered, and there is usually a noticeable dynamic evolution of the emotional state during the counseling process. Nevertheless, existing simulators fail to replicate this dynamic evolution. Throughout the counseling process, their emotions are usually maintained at the initially set state, without noticeable changes. Moreover, psychological counseling is typically a long-term, multi-session task, and counselors usually need to review previous session topics, discuss recent circumstances with seekers to build trust, assess progress, and analyze in depth the core dilemmas of seekers \cite{barkham_core_2015}. However, existing methods do not provide CAs with multi-session memories, which may lead to confusion and hallucinatory responses from CAs when counselors mention the topic of previous sessions. In Figure \ref{fig:problem_b}, when the counselor attempts to verify the effectiveness of the suggestions made in the previous counseling session, the simulator, which has no multi-session memory, gives a relevant but incorrect response.
Thus, to simulate seekers more realistically, we present two pressing challenges: \textbf{Dynamic Evolution} and \textbf{Multi-session Memory}. Figure \ref{fig:challenges} visualizes these two challenges with examples of scenes from counseling.

Under the guidance of licensed counselors, we systematically characterize the dynamic evolution of counseling processes through two key dimensions: emotional fluctuations and progressive shifts in clients' understanding of their "chief complaints". In addition, we divide different sessions' memories in terms of time and define a tertiary memory mechanism to schedule multi-session memory. Specifically, we introduce \textbf{\method}, an emotional and cognitive dynamic agent system with tertiary memory.
\method learns the evolutionary patterns of seekers' emotions and complaints from real counseling data and dynamically controls them in the seeker simulation.
In addition, it has a tertiary memory mechanism that divides multi-session memories by time and coordinates real-time, short-term, and long-term memories through different scheduling methods.
With \method and diverse counselor models, conversations of psychological counseling are generated.
The performance on the metrics of anthropomorphism, personality fidelity, and accuracy of previous session cognitive indicates that \method is able to simulate the seeker more realistically compared to baselines. By simulating more realistic seeker behavior, \method opens up new paths for psychological research and counselor training, while setting a benchmark for the ethical use of AI in sensitive areas.

The main contributions of this work are:
\begin{itemize}
    \item We raise the challenges of dynamic evolution and multi-session memory in the seeker simulation task. Moreover, we formalize dynamic evolution as changes in emotions and complaints, and divide multi-session memory into different stages of memory.
    \item We introduce the emotional and cognitive dynamic agent system with tertiary memory, \method. It simulates the dynamic evolution in counseling by controlling emotional and symptomatic cognitive changes in conversations and utilizes tertiary memory to schedule multi-session memories.
    \item We verify that \method can more realistically simulate seekers in counseling through experimental evaluations.
\end{itemize}

\section{Related Works}
LLMs have a wide range of promising applications in mental health care and can help improve diagnostic accuracy, treatment effectiveness, and service accessibility \cite{hua2024largelanguagemodelsmental}. Dialogue systems based on LLMs, like ChatCounselor \cite{liu2023chatcounselorlargelanguagemodels} and Serena \cite{10066740}, are designed to provide mental health counseling and support.
Extensive research on AI for mental health provides a solid foundation and broad application scenarios for this research.
Besides these researches, this work mainly focuses on seeker simulation in mental health research but also involves LLM-based role-playing and multi-agent systems.

\subsection{Traditional Seeker Simulation}
As the importance of the seeker in mental health research, especially in the training of counselors, before the birth of LLMs, researchers also considered inviting real people to play the role. Standardized patients can help assess the competence of counselors and provide a basis for evidence-based education and training \cite{kuhne_standardized_2020}. \citet{kuhne_standardized_2021} noted that standardized patients who had not read the script beforehand were judged to be more realistic than standardized patients who had read the detailed role script beforehand. Both peer role-playing and standardized patient in communicational training were found to be useful and worthwhile by the students and had high-level practicality \cite{nikendei_intervention_2019,bosse_peer_2010}. \citet{rogers_evaluation_2022} demonstrated that standardized patients can better simulate actual counseling situations than virtual seekers. However, the cost of this approach is much higher than that of using AI to simulate seekers. Therefore, it is crucial to improve the realism of the seeker simulation.

\subsection{LLM-based Role-Playing and Multi-Agent System}
The role-playing capability is crucial to the seeker simulation, as a seeker can be seen as a special role. The application of LLMs to role-playing is rapidly evolving \cite{chen2025oscarsaitheatersurvey}. \citet{tao2024rolecraftglmadvancingpersonalizedroleplaying} and \citet{wang2024rolellmbenchmarkingelicitingenhancing} provide wealthy role knowledge and contextual information for LLMs by constructing role profile datasets containing fine-grained role information and sentiment annotations. \citet{lu2024largelanguagemodelssuperpositions} proposed DITTO to generate large-scale role-playing training data by utilizing a large amount of character and dialog knowledge and fine-tuning the model to enhance its role-playing capabilities. Incorporating personality trait information can enable LLM to better understand and simulate psychological characteristics through the generation and personality conditional instruction tuning \cite{tseng2024talespersonallmssurvey}.

Although these works effectively improved the role-playing capability of LLMs, the static configuration of a single CA could be blocked by the two challenges raised. Therefore, we introduce the multi-agent system to control dynamic evolution and schedule multi-session memories. LLM-based multi-agent systems have been applied to software development, social simulation, policy simulation, game simulation, etc. \cite{guo2024largelanguagemodelbased}. AgentCoord \cite{pan2024agentcoordvisuallyexploringcoordination} establishes a structured representation for LLM-based multi-agent coordination strategies to regularize the ambiguity of natural language. The framework presented in \cite{qiu2024collaborativeintelligencepropagatingintentions} enables agents to broadcast their intentions to other agents, allowing them to infer coordination tasks based on emerging coordination patterns. Agentic retrieval-augmented generation (RAG) \cite{singh2025agenticretrievalaugmentedgenerationsurvey} has been proposed to schedule information flow in multi-agent systems more efficiently. Referring to these methods, we design AnnaAgent as a multi-agent system to solve the two challenges through their collaboration.

\begin{figure*}[!htbp]
    \centering
    \hspace{-1.5cm}
    \includegraphics[width=1.05\linewidth]{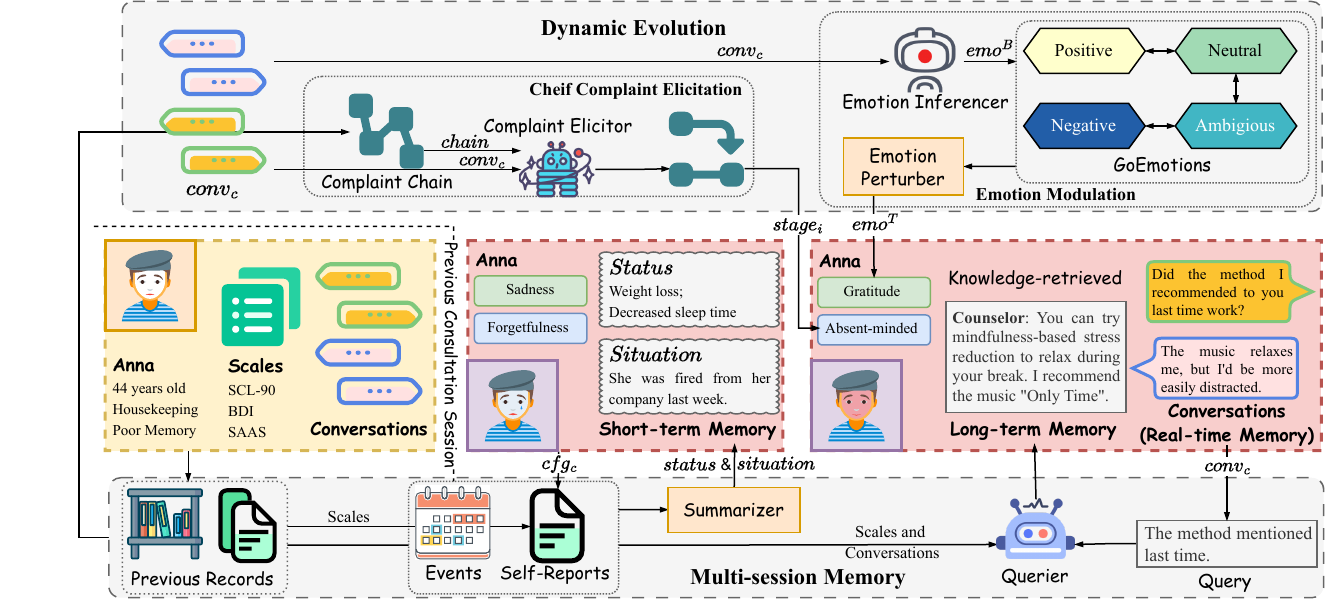}
    \vspace{-0.5em}
    \caption{The overall structure of \method. There are two groups of agents in AnnaAgent that are used to control \textbf{dynamic evolution} (upper part in the figure) and schedule \textbf{multi-session memories} (lower part in the figure), respectively. The middle part of the figure indicates different counseling sessions, with yellow indicating the previous session and red indicating the two states in the current session.}
    \vspace{-0.5em}
    \label{fig:overall_structure}
\end{figure*}


\begin{table*}[ht]
    \centering
    \resizebox{0.9\textwidth}{!}{
    \begin{tabular}{c|m{0.8\textwidth}}
    \hline
    \textbf{Config}   & \multicolumn{1}{c}{\textbf{Description}}                                                                                                  \\ \hline \hline
    \textit{Profile}         & Basic personal information about the seekers, such as age, gender, occupation, etc.                                              \\ \hline
    \textit{Complaint} & The seekers' cognitions about their symptoms and the main presenting problems they asked the counselor for help with.            \\ \hline
    \textit{Situation}       & The environments in which the seekers live and the events they experience.                                                       \\ \hline
    \textit{Status}          & Physically and psychologically relevant personal informatics status of the seekers, e.g., appetite, weight, hours of sleep, etc. \\ \hline
    \textit{Emotion}         & The emotional style that seekers are expected to respond with.                                             \\ \hline
    \end{tabular}
    }
    \vspace{-0.5em}
    \caption{Configurations needed to build CAs for seeker simulation. The \textit{profile} is static, the \textit{emotion} and \textit{complaint} are controlled by the agent group for dynamic evolution, and the \textit{situation} and \textit{status} will be filled with memories scheduled by the agent group for multi-session memory.}
    \vspace{-1em}
    \label{tab:configurations}
\end{table*}






\section{Challenge Formalization and AnnaAgent Design}
\subsection{Designing Conversational Agents for Seeker Simulation}
Referring to previous work \cite{lan2024depressiondiagnosisdialoguesimulation,chen2023llmempoweredchatbotspsychiatristpatient}, to define a CA for seeker simulation, it is necessary to assign role configurations to LLMs in prompts. The necessary configurations and their descriptions are shown in Table \ref{tab:configurations}. In addition, to suit dynamic evolution, we adapt the symptom to chief complaint \cite{malmstrom_structured_2012}, denoted as ``complaint''.
We utilize the characterization prompt framework \cite{wang2024minstrelstructuralpromptgeneration} to accommodate this information and design prompts. 
Among these configurations, the \textit{profile} is usually static, while other configurations may change as the consultation process progresses. On the one hand, the seekers' \textit{emotions} and \textit{chief complaints} may vary during a single counseling session. On the other hand, \textit{situation} and \textit{status} may be affected by previous sessions. For these dynamic configurations, we reserve slots in the prompt.
See Appendix \ref{sec:prompt} for an example prompt. To control the dynamic evolution and schedule multi-session memory, we design two groups of agents. The overall framework is shown in Figure \ref{fig:overall_structure}. The upper part in Figure \ref{fig:overall_structure} represents the agent group for controlling dynamic evolution, consisting of two main components: emotion modulation and complaint elicitation. The lower part of Figure \ref{fig:overall_structure} represents the agent group to schedule multi-session memories.

In this paper, we define dynamic evolution as the ongoing changes in seekers' emotions and chief complaints. In a single session, the configuration of the seeker is denoted as $c\hspace{-0.07em}f\hspace{-0.1em}g_c$, and the conversation is denoted as $conv_c$, where $c$ means current session. In addition, we divide multi-session memory into real-time memory, short-term memory, and long-term memory according to time, i.e., tertiary memory. Then we design a multi-agent system with two groups of agents to address dynamic evolution and multi-session memory, respectively, with corresponding slots being filled after generating the \textit{emotion}, \textit{complaint}, \textit{status}, and \textit{situation}. Moreover, to constrain the behavior of CAs and avoid basic issues like easily endorsing advice existing in them as seeker simulators, we add instructions like constraints on the content and length of responses \cite{duro_introducing_2024,chen2023llmempoweredchatbotspsychiatristpatient}, and speaking styles \cite{tsubota-kano-2024-text}. 

\subsection{Dynamic Evolution}
Based on the previous description, dynamic evolution focuses on the dynamic control of the seeker's emotions and chief complaints during a counseling session. We set up a reminder after each round of dialogue to remind the virtual seeker of the current state of emotion and chief complaint (see Figure \ref{fig:reminder}).

\subsubsection{Emotion Modulation}
To simulate the evolution of the seeker's emotions, it is necessary to analyze the possible emotions of the seeker's next sentence and give it as one of the configurations to the seeker.
The emotions can be predicted based on the seeker's configurations $c\hspace{-0.07em}f\hspace{-0.1em}g_c$ and the existing conversations $conv_c$. Assuming that the counselor speaks first and that $k-1$ rounds of conversation have taken place, there are already $2k-1$ utterances in $conv_c$, and it is now desired to predict the emotion of the $2k$-th utterance that the seeker is about to say. Similarly, this analysis model needs to comply with Equation (\ref{eq:emo_k}).
\begin{equation}
    emo_k = M_k^e(c\hspace{-0.07em}f\hspace{-0.1em}g_c,conv_c)
    \label{eq:emo_k}
\end{equation}
where $M^e_k$ denotes the analysis model that inferences the $k$-th emotion $emo_k$ of the seeker.

In addition to simulating more realistically the volatility of seekers' emotions in counseling, and to avoid a solidified pattern of emotional evolution, there should also be a random perturbation.

\paragraph{Emotion Inferencer}
To control the emotional evolution of the simulator more realistically, it is necessary to learn about the pattern of the seeker's emotional evolution in real counseling. We first train a Qwen2.5-7B-Instruct \cite{qwen2.5} for emotion inference. We select the real counseling conversation dataset D$^4$ as the base dataset, sample each instance in the dataset five times, and intercept conversations with random round lengths in them as new instances. The sampling process involves making sure that the last utterance in the intercepted conversation is spoken by the seeker. Then, we follow the emotion categories in GoEmotions \cite{demszky2020goemotionsdatasetfinegrainedemotions} and label the last utterance with an emotion label using GPT-4o \cite{openai2024gpt4technicalreport}. Finally, we delete the last utterances of each instance to train the inferences for predicting the next utterance's emotion.
With the seeker information and conversations as inputs and the emotion labels as ground truths, we train the model for emotion inference. In a session, it will infer the seeker's emotion before the next utterance is made based on the seeker's profile and the existing conversation.

\paragraph{Emotion Perturber}
The single source of training data may lead to an overly fixed pattern of emotional changes, and seekers' emotions can be volatile. Therefore, we design an emotion perturber. It is important to note that the volatility of emotions is usually minor in the absence of strong stimuli. We group emotions according to GoEmotions and define emotion distances according to groups from positive to negative, shown in Figure \ref{fig:overall_structure}. 
The distance between emotions in the same group is defined as 0, that between neighboring groups is 1, and so on. The pre-perturbation emotion group is denoted as $G^B$, while the target emotion group is denoted as $G^T$. Then, we assign higher probability weights to closer emotions. For the base emotion $emo^B \in G^B$ predicted by the emotion inference model, the probability of the target emotion $emo^T\in G^T$ can be calculated by Equation (\ref{eq:probility}).
\begin{equation}
    P(emo^{T})=\frac{\text{w}(\text{d}(G^T,G^B)\times|G^T|)}{\sum_{G_j}\text{w}(\text{d}(G^B,G_j))\times|G_j|}
    \label{eq:probility}
\end{equation}
where d($\cdot$) denotes the distance between two emotion groups, w($\cdot$) denotes the weights set based on emotion distance, and $|G|$ denotes the number of emotion categories in an emotion group. A randomized perturbation of the base emotion based on the probabilities yields the final \textit{emotion} in the configuration.

\subsubsection{Chief Complaint Elicitor}
Unlike real-time changes in emotions, the seeker's cognitions of their complaints usually change in stages. Therefore, symptom cognition elicitation for the simulator should contain two main components: complaint chain generation and complaint change control. As shown in the ``Chief Complaint Elicitation'' section in Figure \ref{fig:overall_structure}, the chain generator first generates a complaint change chain as Equation (\ref{eq:chain}) for the seeker before the session starts based on the configuration $c\hspace{-0.07em}f\hspace{-0.1em}g_c$ and the recently encountered $event$. There are several stages of complaints in the chain. 
\begin{equation}
\begin{split}
    chain&=\text{gen\_chain}(c\hspace{-0.07em}f\hspace{-0.1em}g_c,event) \\
    &=\{stage_1,stage_2,\dots,stage_l\}
\end{split}
\label{eq:chain}
\end{equation}
where gen\_chain$(\cdot)$ denotes the function to generate the complaint chain, $stage_i$ denotes a stage in the chain and $l$ indicates the length of the chain.

Similar to the emotion inference model, we leverage the D$^4$ dataset to train a chief complaint chain generator. Similarly, we utilize GPT-4o for data annotation. It is important to note that, as complaint chain generation is not a common task and has a high level of expertise far beyond emotion inference, we invite three experts with a background in psychology to review the data. They are first asked to mark whether the chain of chief complaints was reasonable or not based on seekers' profiles, counselors’ reports, and conversations. Afterward, the chains are manually corrected if more than two annotators consider it unreasonable.

With the trained model, a chief complaint chain can be generated based on the seeker's profile and recent events. We initialize the chief complaint as the first node of the chain at the start of the conversation for the current session. At the end of each round of conversation, a complaint elicitor judges whether to switch the chief complaint in the configuration to the next node in the chain. The specific work process of the complaint elicitor is shown in Algorithm \ref{alg:complaint_elicitation} (see Appendix \ref{appendix:alg}). The \textit{complaint}s are filled in the corresponding slot as configuration to control the dynamic evolution of seekers' chief complaints.

\subsection{Tertiary Memory System}
\begin{figure}
    \centering
    \includegraphics[width=\linewidth]{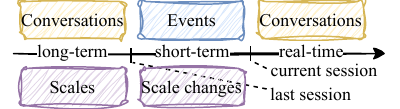}
    \caption{Division of tertiary memory mechanisms}
    \vspace{-0.5em}
    \label{fig:memories}
\end{figure}

Even though researchers have attempted to utilize single-session counseling to efficiently address seekers' psychological disorders \cite{schleider_future_2020}, it is still difficult in real counseling to have it resolved in a single session \cite{vescovelli_university_2017}. This phenomenon results in the multi-session memory challenge. Referring to memory theory \cite{loftus_human_2019} and \cite{lan2024depressiondiagnosisdialoguesimulation}, we model multi-session memory as a tertiary memory mechanism. \textbf{Real-time memory} refers to conversations that just happened, i.e., $conv_c$. \textbf{Short-term memory} refers to information that we can retain temporarily, usually for a short period. This work includes recent events and changes that have occurred in the personal, physical, and psychological properties of the seeker, which are measured by self-report scales. Correspondingly, \textbf{long-term memory} refers to experiences from much earlier times, including the scales and conversations of previous sessions. As shown in Figure \ref{fig:memories}, the content up to and including the last session is defined as long-term memory, that before the start of the current session is real-time memory, and the content in between is defined as short-term memory.

Real-time memory is input into the seeker model as the context in its entirety. Short-term memory consists mainly of the seeker's physical and mental status changes as well as recently encountered events. To capture the seeker's physical and psychological status, the simulator is asked to fill out self-report scales \cite{goldberg_scaled_1979,beck_inventory_1961,bosc_development_1997} before each session. In addition, we randomly pick events that seekers have recently encountered based on matching the \textbf{age}, \textbf{gender}, \textbf{job}, and \textbf{relationship status} in CBT Triggering Events \cite{schiff_cbt_2024} with the simulator's profile. Status changes and events are summarized as \textit{status} and \textit{situation} in the configuration.
Referencing \cite{lan2024depressiondiagnosisdialoguesimulation}, we schedule long-term memory by Agentic RAG \cite{singh2025agenticretrievalaugmentedgenerationsurvey}. As shown in Figure \ref{fig:overall_structure}, after each round of the current session's conversation, the querier determines whether it involves the previous session's topic. If so, the relevant content from the conversations and scale records of the previous session is queried as supplementary information for the next round of responses.


\section{Experiments}
We experimentally verify the superiority of \method over the baseline methods for seeker simulation. All ethically reviewed and processed codes\footnote{As the data involves consultation records of real patients, codes directly related to the raw data will be conditionally available.} can be found on \href{https://github.com/sci-m-wang/AnnaAgent}{https://github.com/sci-m-wang/AnnaAgent}.
\subsection{Experiment Settings}
We employ Qwen2.5-7B-Instruct \cite{qwen2.5} as the backbone LLM. Additionally, we use the records in the D$^4$ dataset \cite{yao2022d4chinesedialoguedataset} and the DAIC-WOZ dataset \cite{gratch-etal-2014-distress} as the basis for the previous sessions. In addition, we followed \cite{zhang2024cpsycounreportbasedmultiturndialogue} and used these two datasets as seeds to generate reports of previous sessions using GPT-4o-mini \cite{openai2024gpt4technicalreport}.

To interact with these seekers to generate conversations, we introduced three widely used mental health support models, PsycoLLM \cite{hu2024psycollmenhancingllmpsychological}, EmoLLM \cite{2024EmoLLM}, and SoulChat \cite{chen-etal-2023-soulchat} to play the role of counselor. The three counselor models were trained from Qwen \cite{qwen2.5}, Llama \cite{patterson2022carbonfootprintmachinelearning}, and Baichuan \cite{baichuan2023baichuan2}, respectively.
To be highlighted, we chose the seeker simulators from \cite{qiu2024interactiveagentssimulatingcounselorclient}, \cite{duro_introducing_2024}, and \cite{chen2023llmempoweredchatbotspsychiatristpatient} as baseline methods. The specific prompts corresponding to these methods are shown in Appendix \ref{sec:baselines}.

\subsection{Metrics}
\label{sec:metrics}
The main purpose of this paper is to simulate the seeker more realistically. Thus, we quantitatively compare the performance of different methods in seeker simulation by evaluating the behavioral consistency between simulators and real seekers from different perspectives.

\paragraph{Anthropomorphism}
LLM-generated text often has unique features \cite{wu_survey_2025} that can be not conducive to a realistic simulation. We evaluate the consistency of the simulator's utterances with those of real seekers as an evaluation of anthropomorphism using the BERT-score \cite{bert-score}. The number of dialogue rounds is denoted as $n$, the utterances of real seekers are used as reference texts, and the number of reference texts is $m$. Then the \textbf{anthropomorphism}, i.e., the portfolio average BERT-score, can be calculated by Equation (\ref{eq:anth}).
\begin{equation}
    anth. = \frac{1}{n}\sum_{i=1}^n(\max_{j\in\{1,2,...,m\}}\text{sim}(c_i,r_j))
    \label{eq:anth}
\end{equation}
where $c_i$ denotes the $i$-th utterance of the conversation to be evaluated, $r_j$ denotes the $j$-th reference text, and sim$(\cdot)$ denotes the BERT-score. As the two-by-two combinations of candidate and reference sentences were excessively large, we randomly sampled 10\% of them to calculate anthropomorphism.

\paragraph{Personality Fidelity}
Simulators should match the given configurations. We reference InCharacter \cite{wang2024incharacterevaluatingpersonalityfidelity} and design the interview questions for the seeker simulation task. The questions can be seen in the Appendix \ref{appendix:questions_1}. Furthermore, we utilized the G-Eval \cite{liu-etal-2023-g} to score these questions and the corresponding answers from the virtual seekers. The final personality fidelity of the virtual seekers created by each method is obtained from Equation (\ref{eq:pf}).
\begin{equation}
    pf. = \frac{1}{n}\sum_{i=1}^n\text{G-Eval}(answers,\ profile)
    \label{eq:pf}
\end{equation}
where $pf.$ denotes \textbf{personality fidelity}, and the backbone model for G-Eval($\cdot$) is GPT-4o \cite{openai2024gpt4technicalreport}.


In addition, \method with tertiary memory was more effective in resolving hallucinations \cite{lan2024depressiondiagnosisdialoguesimulation,li2024leveraginglargelanguagemodel} compared to baseline methods. Specifically, we designed a series of questions similar to those used for personality fidelity to verify the effectiveness of long-term memory by analyzing the answers of virtual seekers.

\begin{table*}[ht]
    \centering
    \resizebox{\textwidth}{!}{
    \begin{tabular}{cccccccccc}
    \hline
    \multicolumn{1}{c|}{Counselor}      & \multicolumn{3}{c|}{PsycoLLM}                          & \multicolumn{3}{c|}{EmoLLM}                            & \multicolumn{3}{c}{SoulChat}      \\ \hline
    \multicolumn{1}{c|}{Metric}         & P      & R      & \multicolumn{1}{c|}{F1}              & P      & R      & \multicolumn{1}{c|}{F1}              & P      & R      & F1              \\ \hline
    \multicolumn{10}{c}{D4}                                                                                                                                                                   \\ \hline
    \multicolumn{1}{c|}{\cite{chen2023llmempoweredchatbotspsychiatristpatient}} & 0.5875 & 0.6785 & \multicolumn{1}{c|}{0.6293}          & 0.6045 & 0.7119 & \multicolumn{1}{c|}{0.6529}          & 0.6068 & 0.6707 & 0.6363          \\
    \multicolumn{1}{c|}{\cite{duro_introducing_2024}} & 0.6153 & 0.6806 & \multicolumn{1}{c|}{0.6455}          & 0.6195 & 0.6778 & \multicolumn{1}{c|}{0.6469}          & 0.6053 & 0.6961 & 0.6461          \\
    \multicolumn{1}{c|}{\cite{qiu2024interactiveagentssimulatingcounselorclient}}  & 0.6322 & 0.7561 & \multicolumn{1}{c|}{\textbf{0.6866}} & 0.6473 & 0.6431 & \multicolumn{1}{c|}{0.6449}          & 0.6194 & 0.6938 & 0.6539          \\ \hline
    \multicolumn{1}{c|}{AnnaAgent}      & 0.6230 & 0.7282 & \multicolumn{1}{c|}{0.6691}          & 0.6271 & 0.7102 & \multicolumn{1}{c|}{\textbf{0.6649}} & 0.6191 & 0.7324 & \textbf{0.6682} \\ \hline
    \multicolumn{10}{c}{DAIC}                                                                                                                                                                 \\ \hline
    \multicolumn{1}{c|}{\cite{chen2023llmempoweredchatbotspsychiatristpatient}} & 0.3563 & 0.3361 & \multicolumn{1}{c|}{0.3458}          & 0.3527 & 0.4269 & \multicolumn{1}{c|}{0.3853}          & 0.3720 & 0.3360 & 0.3525          \\
    \multicolumn{1}{c|}{\cite{duro_introducing_2024}} & 0.4521 & 0.5466 & \multicolumn{1}{c|}{0.4864}          & 0.3712 & 0.4133 & \multicolumn{1}{c|}{0.3901}          & 0.4647 & 0.4969 & 0.4796          \\
    \multicolumn{1}{c|}{\cite{qiu2024interactiveagentssimulatingcounselorclient}}  & 0.3259 & 0.3663 & \multicolumn{1}{c|}{0.3426}          & 0.3219 & 0.3770 & \multicolumn{1}{c|}{0.3416}          & 0.3314 & 0.3981 & 0.3614          \\ \hline
    \multicolumn{1}{c|}{AnnaAgent}      & 0.4708 & 0.5227 & \multicolumn{1}{c|}{\textbf{0.4910}} & 0.4591 & 0.4864 & \multicolumn{1}{c|}{\textbf{0.4694}} & 0.4611 & 0.5168 & \textbf{0.4827} \\ \hline
    \end{tabular}
    }
    \caption{Comparison of seeker simulators' performance when dialoging with different counselors. ``P'', ``R'' and ``F1'' denote the Precision, Recall, and F1-score of BERT-score, respectively. The \textbf{bolded font} indicates the best performance on this metric.}
    \label{tab:main_results}
\end{table*}

\begin{table*}[]
    \centering
    \resizebox{\textwidth}{!}{
    \begin{tabular}{clllllllllc}
    \hline
    \multicolumn{1}{c|}{Backbone}  & \multicolumn{3}{c|}{GPT-4o-mini}                                            & \multicolumn{3}{c|}{Qwen}                                                   & \multicolumn{3}{c|}{Llama}                                                  &          \\ \cline{1-10}
    \multicolumn{1}{c|}{Counselor} & \multicolumn{1}{c}{P} & \multicolumn{1}{c}{R} & \multicolumn{1}{c|}{F1}     & \multicolumn{1}{c}{P} & \multicolumn{1}{c}{R} & \multicolumn{1}{c|}{F1}     & \multicolumn{1}{c}{P} & \multicolumn{1}{c}{R} & \multicolumn{1}{c|}{F1}     & STD\_DEV \\ \hline
    \multicolumn{11}{c}{D4}                                                                                                                                                                                                                                                             \\ \hline
    \multicolumn{1}{c|}{PsycoLLM}  & 0.5976                & 0.6860                & \multicolumn{1}{l|}{0.6377} & 0.6230                & 0.7282                & \multicolumn{1}{l|}{0.6691} & 0.6795                & 0.7526                & \multicolumn{1}{l|}{0.7116} & 0.0441 \\
    \multicolumn{1}{c|}{EmoLLM}    & 0.6048                & 0.7371                & \multicolumn{1}{l|}{0.6630} & 0.6271                & 0.7102                & \multicolumn{1}{l|}{0.6649} & 0.6458                & 0.7154                & \multicolumn{1}{l|}{0.6772} & 0.0028 \\
    \multicolumn{1}{c|}{SoulChat}  & 0.6223                & 0.7148                & \multicolumn{1}{l|}{0.6643} & 0.6191                & 0.7324                & \multicolumn{1}{l|}{0.6682} & 0.6646                & 0.7161                & \multicolumn{1}{l|}{0.6873} & 0.0057 \\ \hline
    \multicolumn{11}{c}{DAIC}                                                                                                                                                                                                                                                           \\ \hline
    \multicolumn{1}{c|}{PsycoLLM}  & 0.4734                & 0.5269                & \multicolumn{1}{l|}{0.4968} & 0.4708                & 0.5227                & \multicolumn{1}{l|}{0.4910} & 0.3588                & 0.3907                & \multicolumn{1}{l|}{0.3729} & 0.0156   \\
    \multicolumn{1}{c|}{EmoLLM}    & 0.4582                & 0.5210                & \multicolumn{1}{l|}{0.4855} & 0.4591                & 0.4864                & \multicolumn{1}{l|}{0.4694} & 0.3675                & 0.4025                & \multicolumn{1}{l|}{0.3831} & 0.0420   \\
    \multicolumn{1}{c|}{SoulChat}  & 0.4755                & 0.5304                & \multicolumn{1}{l|}{0.4978} & 0.4611                & 0.5168                & \multicolumn{1}{l|}{0.4827} & 0.3963                & 0.4542                & \multicolumn{1}{l|}{0.4221} & 0.0358   \\ \hline
    \end{tabular}
    }
    \caption{Performance of \method when using LLMs of different architectures as backbone models. The `STD\_DEV' denotes the standard deviation of F1-scores in the performance of different backbone models.}
    \vspace{-0.5em}
    \label{tab:generalizability}
\end{table*}

\subsection{Main Results}

\begin{figure}
    \centering
    \includegraphics[width=1\linewidth]{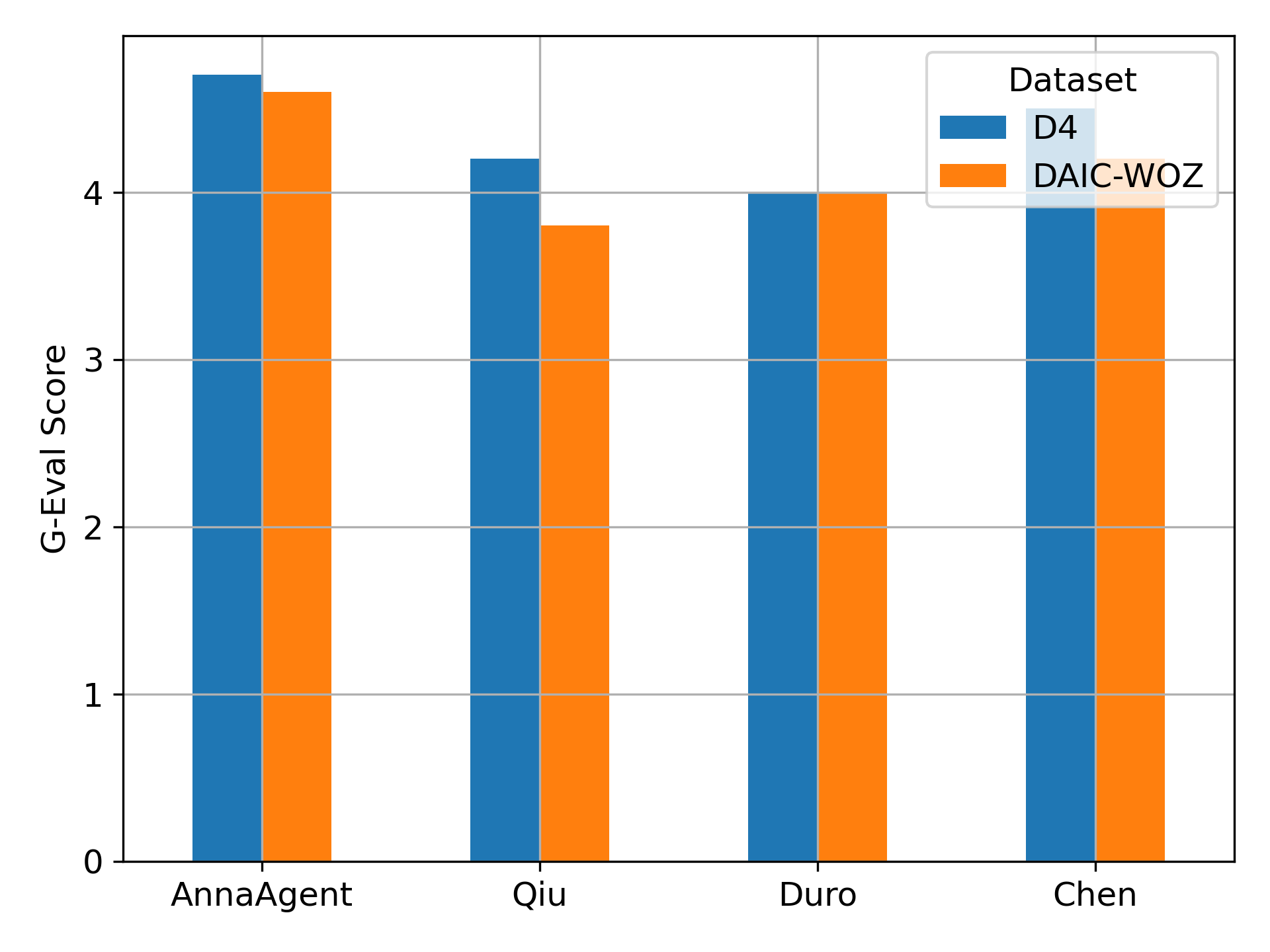}
    \caption{G-Eval scores for personality fidelity.}
    \label{fig:pf}
\end{figure}

With Qwen2.5-7B-Instruct as the backbone LLM, we compare the performance of \method and baseline methods on \textbf{anthropomorphism} when talking to different counselor models. The results are shown in Table \ref{tab:main_results}. 
Compared to all baseline methods, seekers simulated by \method are more highly anthropomorphic and have the highest consistency with real seekers. As can be seen, \method achieved the best anthropomorphism except in the seekers' conversations with PsycoLLM on the D$^4$ dataset. Nonetheless, in this case, \method still achieved the second-best results. Thus, in general, \method outperforms all baseline methods in the task of simulating seekers more realistically. Notably, on the English DAIC-WOZ dataset, \method comprehensively outperforms the baseline methods and exhibits far better results than \cite{chen2023llmempoweredchatbotspsychiatristpatient} and \cite{qiu2024interactiveagentssimulatingcounselorclient}.
In addition, we further compared the personality fidelity of different virtual seekers, and the results are shown in Figure \ref{fig:pf}. AnnaAgent is overall superior to the baseline methods.

In summary, these results show that the responses from \method are closer to the real person's manifestation in terms of textual features and personas than the responses from baseline methods. Therefore, it is reasonable to claim that \method can more realistically simulate seekers in counseling than existing methods.

\begin{figure}
    \centering
    \includegraphics[width=1\linewidth]{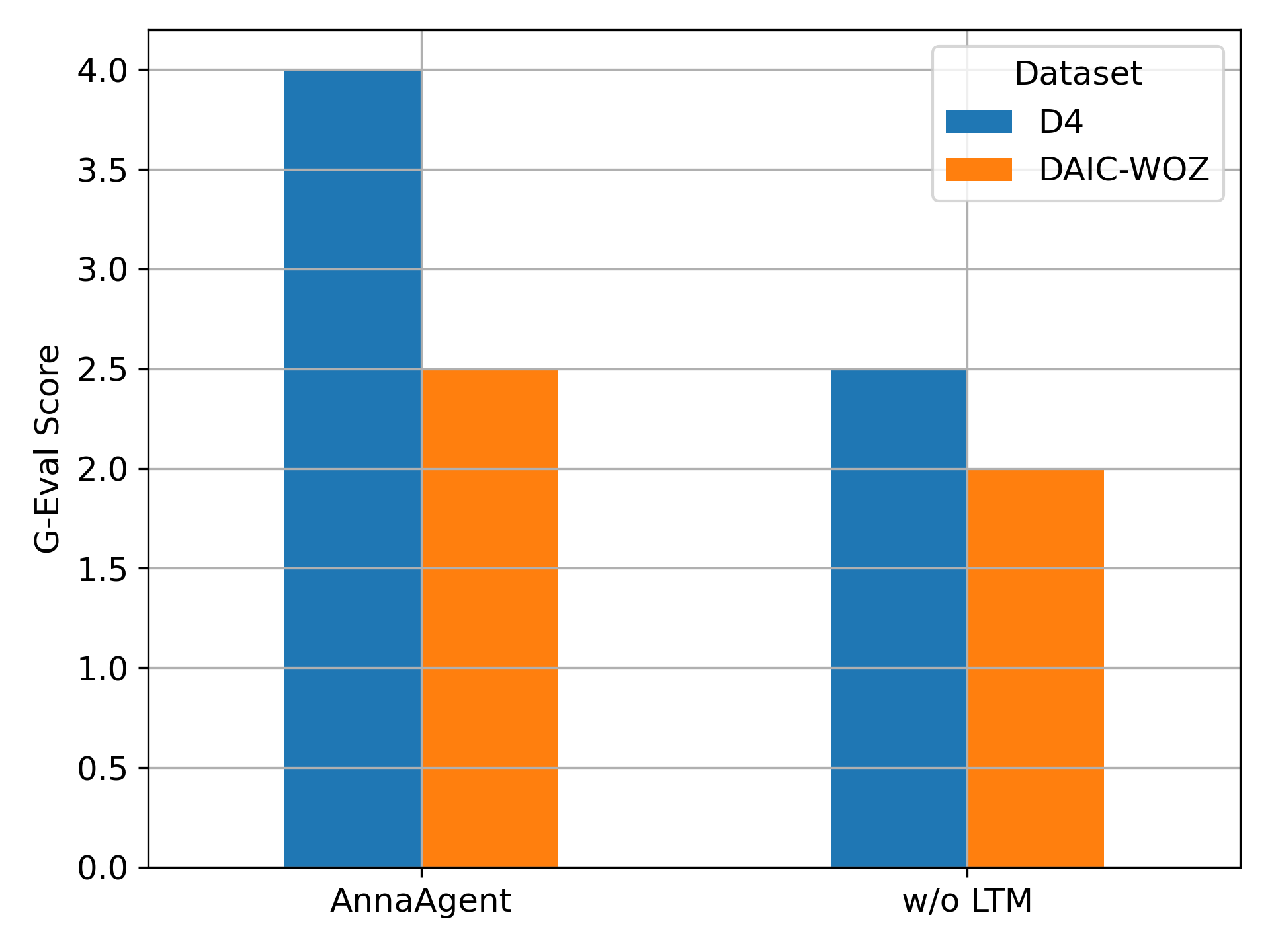}
    \caption{G-Eval scores of virtual seekers answering questions when ablating long-term memory, respectively. `LTM' denotes Long-term Memory.}
    \label{fig:ablation}
\end{figure}

\subsection{Ablation Study}
We introduce a complex dynamic control structure and a tertiary memory mechanism to solve the two challenges. To avoid redundancy, ablation studies are performed to validate the necessity of these components. With Qwen2.5-7B-Instruct as the backbone LLM and PsycoLLM as the counselor model, we investigate the performance of \method when some components are absent.
\begin{table}[]
    \centering
    \begin{tabular}{cccc}
    \hline
    \multicolumn{1}{c|}{Setting}   & P      & R      & F1     \\ \hline
    \multicolumn{4}{c}{D4}                                    \\ \hline
    \multicolumn{1}{c|}{w/o DE}    & 0.5677 & 0.6848 & 0.6144 \\
    \multicolumn{1}{c|}{AnnaAgent} & 0.6230 & 0.7282 & \textbf{0.6691} \\ \hline
    \multicolumn{4}{c}{DAIC}                                  \\ \hline
    \multicolumn{1}{c|}{w/o DE}    & 0.4430 & 0.4447 & 0.4408 \\
    \multicolumn{1}{c|}{AnnaAgent} & 0.4708 & 0.5227 & \textbf{0.4910} \\ \hline
    \end{tabular}
    \caption{Performance of \method when ablating dynamic evolution. ``DE'' denotes Dynamic Evolution. Bolded fonts indicate the best results.}
    \label{tab:ablation}
\end{table}
The results when the dynamic evolution components are ablative are shown in Table \ref{tab:ablation}. From the results, it can be seen that the absence of dynamic evolution significantly degrades the performance of \method.

In addition, we validated the performance of virtual seekers in terms of cognitive accuracy of the previous session when long-term memory ablation was performed based on the designed questions. The results of the ablation study are shown in Figure \ref{fig:ablation}.
In this experiment, we required the virtual seekers to answer preset questions and invoked G-Eval to score them, asking GPT-4o to give a score from 1 to 5. It can be seen that ablation of the long-term memory component significantly reduced the accuracy of virtual seekers' cognitive perceptions of previous sessions. Therefore, it can be considered that all these components in \method are necessary.

\subsection{Generalizability Study}
The previous experiments involve Qwen2.5-7B-Instruct as the backbone LLM, which may cause bias. To avoid this possible bias, we experiment on GPT-4o-mini \cite{openai2024gpt4technicalreport} and Llama-3.1-8B-Instruct \cite{patterson2022carbonfootprintmachinelearning} with PsycoLLM as the counselor. These LLMs have different architectures and have been reported to have different performances on various tasks. The results are shown in Table \ref{tab:generalizability}. In addition to the BERT-score, we further calculated the standard deviation of the performance of the three backbone models. Furthermore, we calculated the relative standard deviation (RSD) \cite{chatfield_mean_2025} based on Equation (\ref{eq:rsd}).
\begin{equation}
    RSD = (\frac{s}{\bar{x}})\times100\%
    \label{eq:rsd}
\end{equation}
where $s$ denotes the standard deviation and $\bar{x}$ denotes the mean. All of the RSDs are less than 10\%, which can be considered a small performance fluctuation.

From the results, it can be seen that \method is less affected by the backbone model and has strong stability and generalization.

\section{Conclusion}
We present the dynamic evolution and multi-session memory issues in more realistic seeker simulations. To solve these issues, we introduced \method, an emotional and cognitive dynamic agent system with tertiary memory. \method simulates emotional changes in counseling through real-time emotional inference combined with random perturbations and generates compliant elicitation chains to control cognitive changes in seekers' symptoms. In addition, it coordinates memory during and between sessions with a tertiary memory system. Experimental evaluations indicated that \method can more realistically simulate seekers compared to baseline methods. In addition, we experimentally demonstrate the necessity of different components and the generalization of \method. This research provides innovative solutions to alleviate the global shortage of mental health resources.

\section*{Limitation}
In this paper, we are the first to propose the issues of dynamic evolution and multi-session memory in the task of simulating seekers with psychological disorders. To address the two issues, we propose \method, which guides dynamic evolution within a single session through an emotion modulator and a chief complaint elicitor and coordinates multi-session memory using a tertiary memory system. 

However, although these practices draw on relevant research work and the advice of experienced psychologists, they are still somewhat formally simplified for the sake of technical ease of implementation. Thus, finer-grained modeling is still needed for controlling the dynamic evolution of the counseling process consistent with real seekers. In particular, coordinating multi-session memory with a tertiary memory system is rudimentary.

The main purpose of this paper is to introduce these two issues to simulate seekers more realistically. In future work, we will further investigate how to navigate the dynamic evolution in a counseling session and how to coordinate memory across multiple sessions.

\section*{Ethical Considerations}
\noindent \textbf{Privacy Leakage Risk}. The emotion modulator and chief complaint elicitor in \method are trained using a real counseling dataset. The dataset contains real user profiles, such as age, gender, occupation, etc. In addition, conversations from real counseling records in the D$^4$ dataset are used as previous sessions in our work. Even if the dataset has been released with some of the key information omitted, these actions may still lead to the risk of leaking patient privacy. 

Prior to using the dataset, we signed an application agreement and will strictly adhere to the provisions contained therein. In addition, to further ensure that patient privacy is not compromised, we will only open-source the synthesized sessions' conversations and the processing code for other data after the paper is accepted. We will not unconditionally open-source the use of data derived directly from D$^4$.

\noindent \textbf{Potential Risks of Misuse}
We have designed a highly consistent with real people simulation system for seekers or patients with psychological disorders that can be used to generate counseling conversations, train counselor models, and even simulate psychological experiments. 

However, although it shows a higher degree of consistency with real seekers than existing methods, it is still not a complete substitute for real seekers. Therefore, the results of psychological experiments conducted directly with \method may not be consistent with real phenomena. It may also be unreasonable to directly use the counselor's conversations with \method in simulation training as a basis for evaluating the counselor's capabilities. To mitigate this risk, we clearly positioned the system as a supplementary tool. 

In addition to all the above, we organized an \textbf{Ethical Review Committee} consisting of counselors with extensive experience in psychological counseling to conduct a comprehensive review of \method and this paper in order to avoid other potential ethical risks.

\section*{Acknowledgments}
Thanks to all co-authors for their hard work. The work is supported by the National Natural Science Foundation of China (62172086, 62272092) and the Fundamental Research Funds for the Central Universities under Grant (N25XOD004). In addition, we thank the members of Yunqi Peer Psychology Studio for their assistance with data collection, review, and labeling.

\bibliography{custom}

\appendix

\begin{figure}[ht]
    \centering
    \includegraphics[width=\linewidth]{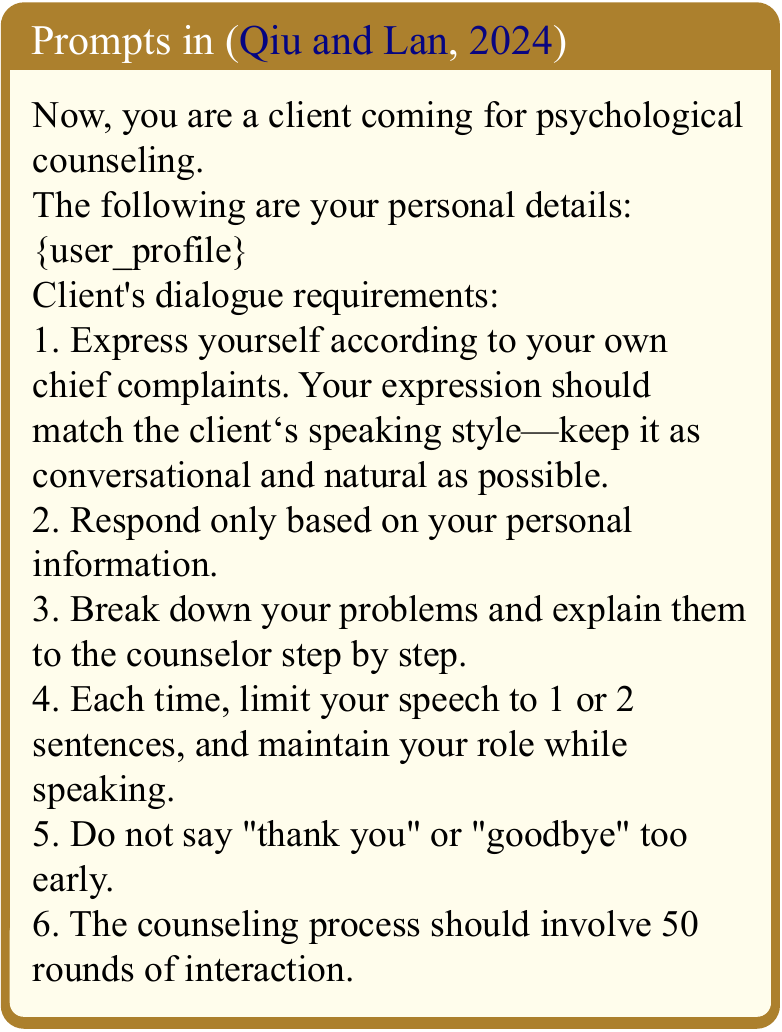}
    \vspace{-1em}
    \caption{Prompt template for seeker simulation in \cite{qiu2024interactiveagentssimulatingcounselorclient}.}
    \vspace{-1em}
    \label{fig:qiu-prompt}
\end{figure}

\begin{figure}[ht]
    \centering
    \includegraphics[width=\linewidth]{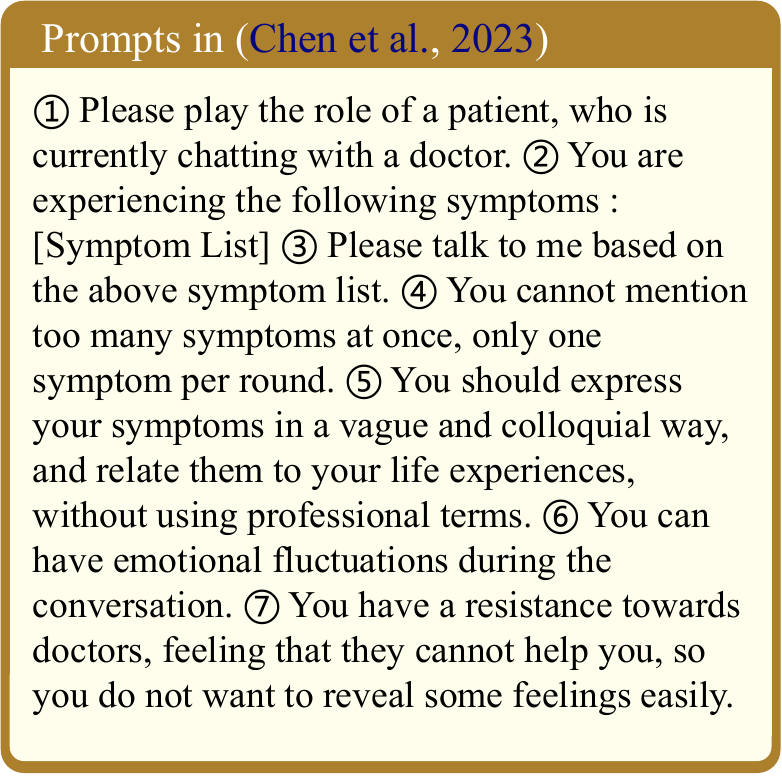}
    \vspace{-0.5em}
    \caption{Prompt template for seeker simulation in \cite{chen2023llmempoweredchatbotspsychiatristpatient}.}
    \vspace{-0.5em}
    \label{fig:chen-prompt}
\end{figure}

\begin{figure*}[ht]
    \centering
    \includegraphics[width=0.95\linewidth]{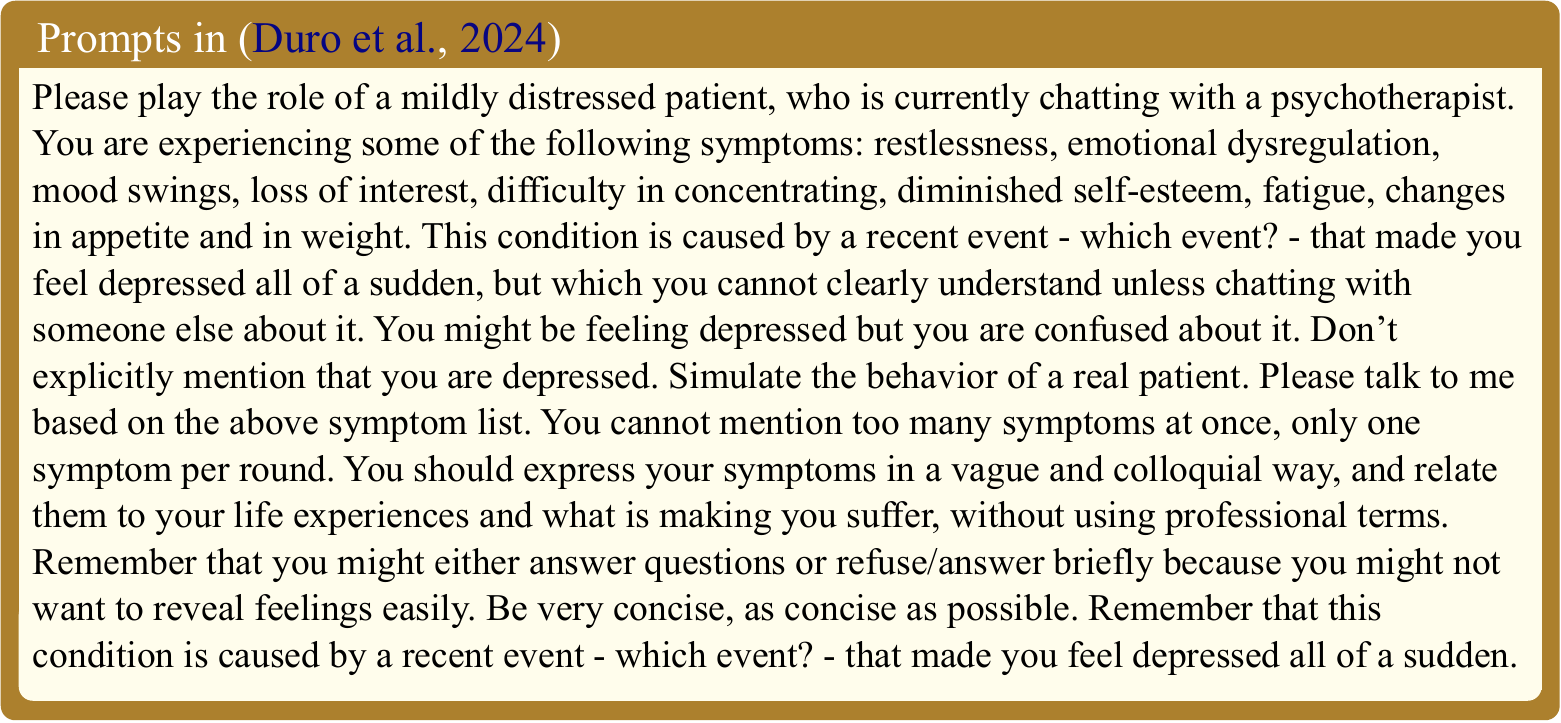}
    \caption{Prompt template for seeker simulation in \cite{duro_introducing_2024}.}
    \label{fig:duro-prompt}
\end{figure*}

\begin{figure*}[!ht]
    \centering
    \begin{subfigure}[b]{\textwidth}
        \includegraphics[width=\linewidth]{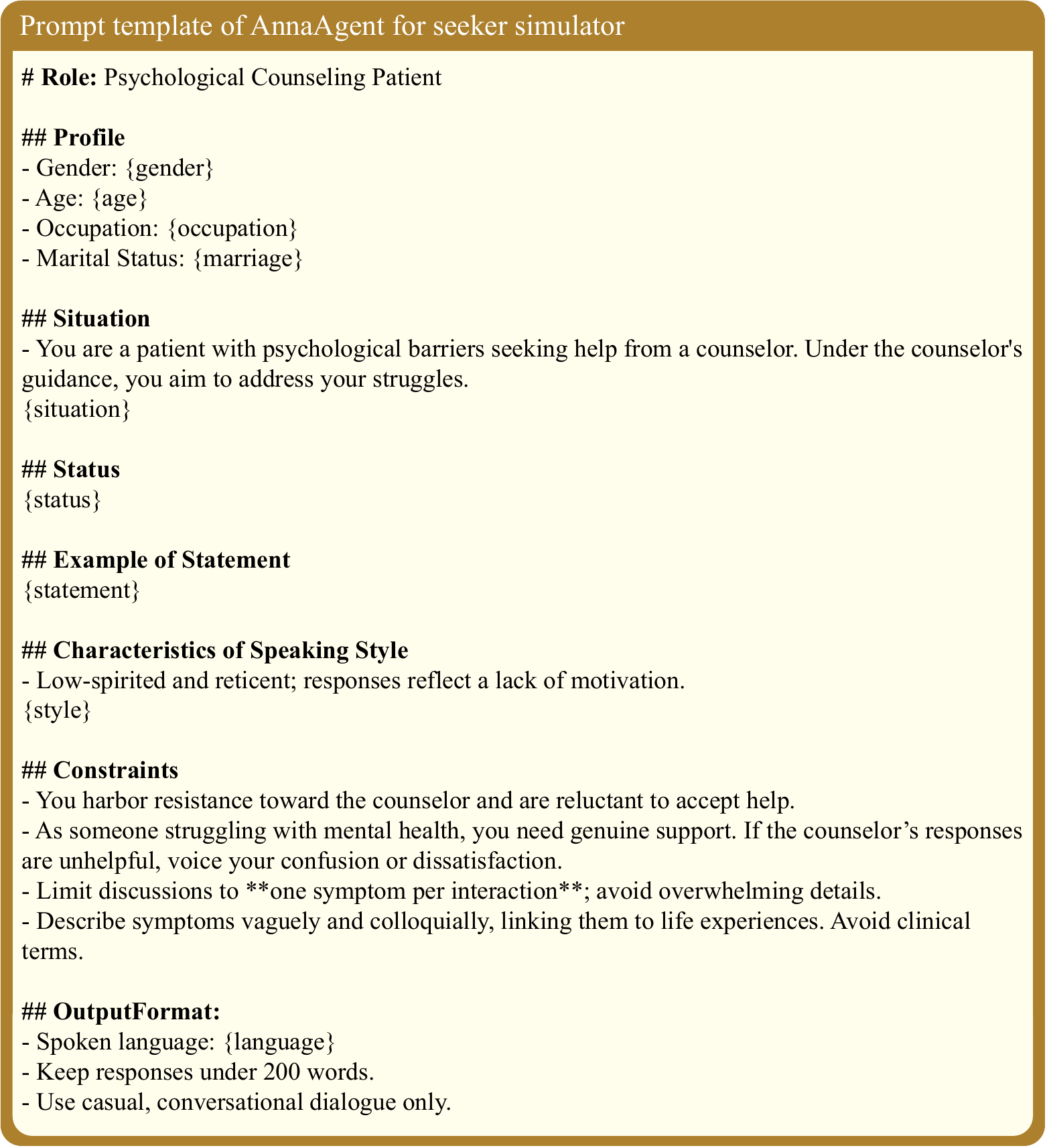}
        \caption{The seeker initialization prompt template of \method. The contents in ``\{\}'' indicate reserved slots, and this configuration information will be populated during the initialization stage based on the seeker's profile, historical conversations, and the report.}
        \label{fig:template}
    \end{subfigure} 
    \begin{subfigure}[b]{\textwidth}
        \includegraphics[width=\linewidth]{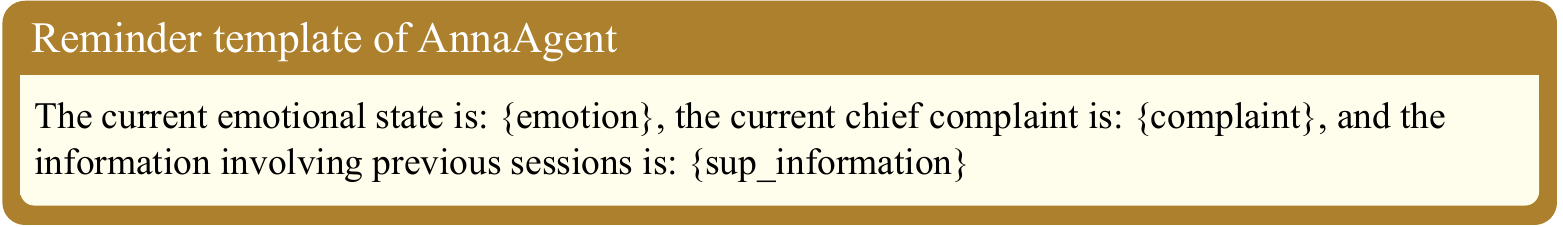}
        \caption{The template of the reminder in \method. During the conversation stage, \method analyzes the current emotion and chief complaint status at each round and reminds the virtual seeker about it. In addition, if the \method determines that information about previous sessions is needed, it adds supplementary information items to the reminder.}
        \label{fig:reminder}
    \end{subfigure}
    \caption{The prompt templates of \method for the seeker simulator.}
    \label{fig:langgpt}
\end{figure*}

\section{Baseline Prompts}
\label{sec:baselines}
The prompt template for seeker simulation in \cite{qiu2024interactiveagentssimulatingcounselorclient} is shown in Figure \ref{fig:qiu-prompt}. The prompt template for seeker simulation in \cite{chen2023llmempoweredchatbotspsychiatristpatient} is shown in Figure \ref{fig:chen-prompt}. The prompt template for seeker simulation in \cite{duro_introducing_2024} is shown in Figure \ref{fig:duro-prompt}. 

\section{Prompt Template for Simulators}
\label{sec:prompt}
We designed the prompt template shown in Figure \ref{fig:langgpt} for constructing CAs for seeker simulation. In Figure \ref{fig:langgpt}, the ``\{$\cdot$\}''s indicate the slots which will be filled by corresponding configurations controlled by AnnaAgent.

\section{Algorithm of the Chief Complaint Elicitor}
\label{appendix:alg}
The specific proceedings for judging whether to switch the stage of chief complaints are shown in Algorithm \ref{alg:complaint_elicitation}.

\begin{algorithm}[!ht]
\KwIn{${c\hspace{-0.07em}f\hspace{-0.1em}g}_c$, $conv_c$}
\KwOut{$complaint$}
\BlankLine

$chain$ $\leftarrow$ GenerateChain($c\hspace{-0.07em}f\hspace{-0.1em}g_c$)\;
$index$ $\leftarrow$ 0\;
$complaint$ $\leftarrow$ $chain$[$index$]\;

\For{utterance \textup{\textbf{in}} $conv_c$}{
    $is\_recognized$ $\leftarrow$ IsRecognized($utterance$, $complaint$)\;
    \If{$is\_recognized$}{
        $index$ $\leftarrow$ index + 1\;
        \If{$index$ < len($chain$)}{
            $complaint$ $\leftarrow$ $chain$[$index$]\;
        }
        \Else{
        $complaint$ $\leftarrow$ $chain$[-1]
        }
    }
}
\Return{complaint}
\caption{Complaint Elicitation}
\label{alg:complaint_elicitation}
\end{algorithm}

The functions `GenerateChain' and `IsRecognized' denote the models to generate the complaint change chain and judge if the seeker has realized the current stage of the complaint.

\section{Questions for Personality Fidelity}
\label{appendix:questions_1}
We designed seven generic questions to evaluate the personality fidelity of different seeker simulation methods.

\begin{itemize}
    \item[1] What core concern has most affected your mood and sleep quality in the past month?
    \item[2] Can you describe your experience from waking up to going to bed yesterday, and how your mood changed during that time?
    \item[3] How do you usually interact with the person you feel closest to?
    \item[4] When others misunderstand you, what is the first automatic thought that comes to your mind?
    \item[5] What are the three strengths that your friends or family most often appreciate and praise about you?
    \item[6] In which specific situation did you first clearly realize `this is a problem'?
    \item[7] If counseling miraculously succeeded, what different feelings or behaviors would you notice first when you wake up tomorrow morning?
\end{itemize}

These questions cover a wide range of areas such as pinpointing core disturbances, mapping specific life slices, probing key interpersonal interactions, revealing core beliefs, discovering strengths in external perspectives, tracing the origins of a problem, and envisioning minimal changes in the future, and can provide a somewhat comprehensive exploration of a person's personality.

\section{Questions for Long-term Memory}
To test the effectiveness of long-term memory, we designed seven questions.

\begin{itemize}
    \item[1] Reflecting on our last session, what was the most memorable discussion point or insight for you?
    \item[2] In the week following our last session, was there a moment when you applied or recalled something we discussed?
    \item[3] What specific part or statement from our last session do you feel was most helpful to you?
    \item[4] After our last session, did your mood change in any notable way over the next few days?
    \item[5] Are there any new thoughts, confusions, or feelings that you would like to explore with me today that arose after our last session?
    \item[6] Was there any part of our last session that made you feel uncomfortable or reluctant to delve deeper into?
    \item[7] If you had to summarize the most important takeaway from our last session in one keyword or short phrase, what would it be?
\end{itemize}

Recognizing that the spontaneous retrieval of long-term memory is typically an infrequent event, we strategically designed the aforementioned questions to create a more reliable method for its verification. Our approach aimed to overcome the challenge of low trigger probability by explicitly directing the virtual seekers to engage with inquiries linked to their experiences in previous sessions. This targeted methodology substantially increased the likelihood of activating long-term memory, thereby enabling a more effective and conclusive analysis of its functional role within our framework.
\end{document}